\documentclass[10pt,twocolumn,letterpaper]{article}

\usepackage{iccv}
\usepackage{times}
\usepackage{epsfig}
\usepackage{graphicx}
\usepackage{amsmath}
\usepackage{amssymb}
\usepackage{color}
\usepackage{booktabs}       % professional-quality tables
\usepackage{tabularx}
\usepackage{array}
\usepackage{float}
\usepackage{multirow}
\usepackage{graphicx}% http://ctan.org/pkg/graphicx
\usepackage{caption}
\usepackage{makecell} % for line breaks in table

\definecolor{MyDarkBlue}{rgb}{0,0.08,0.5}
\definecolor{MyDarkGreen}{rgb}{0.02,0.30,0.02}
\definecolor{MyDarkRed}{rgb}{0.7,0.02,0.02}
\definecolor{MyDarkOrange}{rgb}{0.40,0.2,0.02}
\definecolor{MyIndigo}{RGB}{111,0,255}

\newcolumntype{Y}{>{\centering\arraybackslash}X}

\usepackage[pagebackref=true,breaklinks=true,letterpaper=true,colorlinks,bookmarks=false]{hyperref}

\iccvfinalcopy % *** Uncomment this line for the final submission

\def\iccvPaperID{1976} % *** Enter the ICCV Paper ID here

\ificcvfinal\fi

\title{Generating Training Data for Denoising \\Real RGB Images via Camera
Pipeline Simulation}

\author{ \normalsize \bf Ronnachai Jaroensri
  $\quad$Camille Biscarrat$\quad$Miika Aittala
  $\quad$Fr{\'e}do Durand\\ \normalsize
  MIT CSAIL\\
  \{tiam,cjbisc,miika,fredo\}@csail.mit.edu
}

\begin{document}

\twocolumn[{%
\renewcommand\twocolumn[1][]{#1}%
\maketitle
\begin{center}
    \makebox[\textwidth][c]{
    \setlength\tabcolsep{3pt}
    \begin{tabular}{cccc}
     \includegraphics[width=0.25\linewidth]{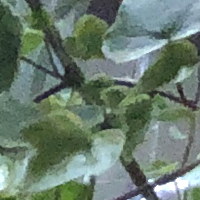} &
     \includegraphics[width=0.25\linewidth]{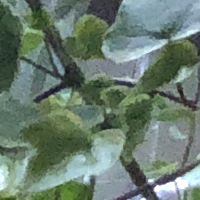} &
     \includegraphics[width=0.25\linewidth]{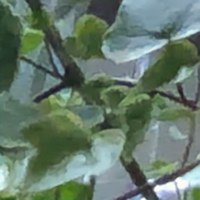} &
     \includegraphics[width=0.25\linewidth]{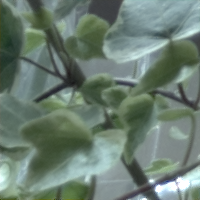} \\
        \thead{(a) Input real noisy JPEG \\ (32.4dB)} & 
        \thead{(b) N3Net\cite{plotz2018neural} trained \\ with AWGN (32.3dB)} &
        \thead{(c) N3Net\cite{plotz2018neural} trained \\ with our pipeline (35.2dB)} &
        \thead{(d) Ground truth \\ (demosaicked)} \\

 \end{tabular}
    }
    \captionof{figure}{(a) Real noise in cellphone-processed JPEG pictures is very
    different from uncorrelated Gaussian noise widely assumed (see Fig.
    \ref{fig:awgn_vs_real_noise}). (b) A blind denoiser
    trained on additive white Gaussian noise (AWGN) is unable to recognize the
    noise pattern resulting and denoise the image. (c) In contrast, the network trained on
    realistic noise model generated by our pipeline was able to denoise
    properly, resulting in over 3dB improvement.}

 \label{fig:teaser}
\end{center}%
}]
%%%%%%%%% TITLE

%\thispagestyle{empty}
\begin{abstract}
    \vskip -5mm
    Image reconstruction techniques such as denoising often need to be applied
    to the RGB output of cameras and cellphones. Unfortunately, the commonly used
    additive white noise (AWGN) models 
    do not accurately reproduce the noise and the degradation
    encountered on these inputs. This is particularly important for learning-based
    techniques, because the mismatch between training and real world data will hurt
    their generalization.
    This paper aims to accurately simulate the degradation and noise transformation performed by camera
    pipelines. This allows us to generate realistic degradation in RGB
    images that can be used to train machine learning models.
    We use our simulation to
    study the importance of noise modeling for learning-based denoising. Our
    study shows that a realistic noise model is required for learning to denoise real JPEG images.
    A neural network trained on realistic noise outperforms the one trained with AWGN by 3 dB.
    An ablation study of our pipeline shows that simulating denoising and demosaicking is important to this
    improvement and that realistic demosaicking algorithms, which have been rarely considered, is needed.
    We believe this simulation will also be useful for other image
    reconstruction tasks, and we will distribute our code publicly.
\end{abstract}

%\keywords{Image restoration, image processing, cellphone, DSLR, denosing, deblurring}

\section{Introduction}

\begin{figure}[t]
\centering
\begin{tabular}{cc}
    AWGN  & \thead{Real Noise from \\ Pixel XL phone} \\
     \includegraphics[width=0.45\linewidth]{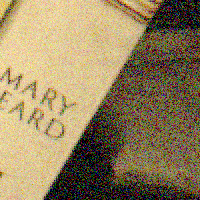}
     &
     \includegraphics[width=0.45\linewidth]{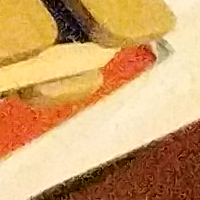} \\
 \end{tabular}

    \caption{
        AWGN noise (left) and real noise from Pixel XL phones (right).
        The noise in real images is processed extensively by the camera
        pipeline. For this particular camera, the artifact is long-grained
        (right) and very different from the fine chroma pattern of AWGN (left).
    }
       \vskip -5mm
\label{fig:awgn_vs_real_noise}
\end{figure}

Most image reconstruction techniques such as denoising operate on RGB images,
either JPEGs directly from a camera or RAW files that have been demosaicked
later. In this paper, we show that the simple additive Gaussian noise (AWGN)
usually used in the literature
\cite{zhang2018ffdnet,lehtinen2018noise2noise,zhang2017beyond} does not
accurately model the artifacts observed in real image.
This is especially true when
working from JPEG images, which undergo a long pipeline that includes operations
such as demosaicking, denoising, and compression that dramatically transform the
noise (see Fig. \ref{fig:awgn_vs_real_noise}). 
The mismatch of noise profiles can
have a strong adverse effect on performance, especially for learning-based
approaches.

Several works have shown that image reconstruction tasks can benefit from better
noise modeling \cite{aittala2018burst,dong2018learning,brooks2018unprocessing,guo2018toward}.
However, most noise and degradation models found in the literature remain simplistic. For
example, most works do not consider demosaicking artifacts, and the ones that do typically
use bilinear demosaicking
\cite{brooks2018unprocessing,guo2018toward,plotz2017benchmarking}, which is rarely used in real
consumer cameras \cite{heide2014flexisp,hasinoff2016burst}.

In this paper, we propose a camera simulation pipeline that can be used to
realistically simulate camera processing of images. We implement over 40 individual modules that can
be custom-built into a camera pipeline. While not exhaustive, they cover a good range of
typical camera modules such as tonemapping, demosaicking, and denoising. From
these modules, we build a pipeline capable of processing RAW images, with some
manual tuning, into
visually similar RGB images that some cellphone cameras produce.

We believe this pipeline can be used to generate data for many low-level vision
tasks. To demonstrate, we use our pipeline to study the
importance of noise modeling in supervised denoising of JPEG images.
We generate different versions of a dataset,
where the pipeline processing parameters and, therefore, the noise characteristics vary. We train state-of-the-art denoising
convolutional neural networks (CNN) \cite{plotz2018neural} on these datasets and measure their performance on real processed
images. We show that networks trained with our realistic pipeline outperformed ones that
are trained on AWGN by roughly 3 dB on real test images (see Fig. \ref{fig:teaser} and
Fig. \ref{fig:awgn_v_pipeline}).

To understand how our pipeline contributes to this improvement, we train
additional networks with different combinations of simulation components. We find that
performance drops markedly if denoising and/or demosaicking components are removed.
Furthermore, the choice of demosaicking algorithm is also important. 
Using bilinear demosaicking in camera simulation pipelines
\cite{brooks2018unprocessing,guo2018toward,plotz2017benchmarking} leads to less
effective denoising compared to using edge-aware methods such as the Adaptive
Homogeneity-Directed (AHD) algorithm
\cite{hirakawa2005adaptive}.

While we are not the first to propose a camera simulation
\cite{karaimer2016software,farrell2003simulation}, our main contribution is to
integrate it into the learning pipeline and use it to show the importance of
realistic simulation for learning-based image restoration tasks. We summarize
our contributions as follow:
\begin{enumerate}
    \item We propose a camera simulator that is expressive enough to simulate
        processing of real cameras. We believe
        that this simulator will be useful for many learning-based image restoration tasks.
    \item Using this simulator, we study the importance of realistic noise
        modeling for denoising real images. We show that:
    \begin{enumerate}
        \item A realistic noise model is beneficial for denoising real JPEG
            images and leads to superior performance compared to AWGN.
        \item Denoising filters and demosaicking are the most important
            components for simulating realistic noise.
        \item Realistic demosaicking algorithm is important and leads to
            improvement over bilinear demosaicking commonly used in camera simulation pipelines
            \cite{brooks2018unprocessing,guo2018toward,plotz2017benchmarking}.
    \end{enumerate}
\end{enumerate}

All code and evaluation datasets will be released publicly.

\section{Related Work}

Classical denoising techniques often create probabilistic models of the noise and
signal and use this model to derive a denoising algorithm. Wavelet coring is 
based on the observation that noise is usually smaller than the image signal,
resulting in smaller wavelet coefficients that can be suppressed
\cite{simoncelli1996noise,portilla2003image,donoho1994ideal}. The current
state-of-the-art classical denoising method remains BM3D \cite{dabov2006image}. The
algorithm performs non-local matching within the image and average these matched
blocks together. These methods typically assume an AWGN model in order to simplify
their modeling effort.

With the growing popularity of CNNs \cite{krizhevsky2012imagenet},
learning-based denoising is becoming prevalent. DnCNN \cite{zhang2017beyond} uses CNNs to
predict a residual map that corrects noisy images. N3Net \cite{plotz2018neural}
formulates a differentiable version of nearest neighbor search to
further improve DnCNN. FFDNet \cite{zhang2018ffdnet} attempts to address
spatially varying noise by appending noise level maps to the input of DnCNN.
Despite many improvements, these works perform very similarly, with often
less than 0.5 dB difference. Moreover, they assume spatially uncorrelated noise,
which is not true for real noisy JPEG images.

Many works in image processing are recognizing the important of noise modeling.
\cite{gharbi2016deep,heide2014flexisp} jointly model noise with their
demosaicking task and found it to improve their performance.
\cite{aittala2018burst} uses an adhoc noise model that simulates spatial
correlation of noise. They found this to significantly boost the quality of
their deblurring results.

Recent denoising work proposes simulating camera pipelines.
\cite{brooks2018unprocessing} unprocesses JPEG images to get RAW representation 
and focus on RAW-to-RAW denoising. Very related to our work is
\cite{guo2018toward} who also uses simulated camera pipeline to supplement real
training data. However, these works tend to assume a limited camera pipeline and
do not evaluate on real processed images.
Our work follows in the same spirit, though
we aim to accurately model realistic camera pipelines, and evaluate our results
with real images.

\section{Camera Simulation Pipeline}
\label{sec:cam_sys}

\begin{figure}[t]
\centering
\includegraphics[width=1\linewidth]{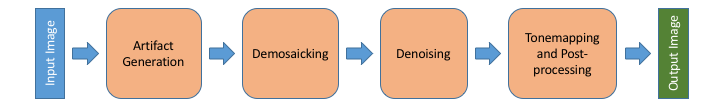}\caption{
Our camera pipeline consists of four main stages: artifact generation,
    demosaicking, denoising, and post-processing.
}
   \vskip -5mm
\label{fig:campipe}
\end{figure}

Our camera pipeline is designed to mirror typical camera processing
stages. We build our pipeline from individual modules that are easily extensible.
Additionally, we also include an
\emph{artifact generation} stage that simulates the degradation of the image signal,
by introducing artifacts such as noise and motion blur.

Fig. \ref{fig:campipe} shows the 4 main stages of our pipeline: artifact simulation, demosaicking, denoising,
and postprocessing. In sum, we have over 70 parameters that
control the behavior of our pipeline. We describe each stage as follows.

\textbf{Artifact Generation.} The first stage of our pipeline is the physical artifact simulators. It aims to
simulate the physical degradation process that happens before the sensor.
It includes motion blur, chromatic aberration, multiplicative exposure
adjustment, and noise.

Noise at the sensor is largely uncorrelated and zero-mean. So we only simulate
spatially uncorrelated noise here. 
Because photon noise is Poisson
in nature and the sensor read noise is Gaussian, we provide both additive and
multiplicative noise to simulate the two effects. 
We optionally mosaick input images before adding noise if the user
wishes to simulate the Bayer pattern, which will then be demosaicked and processed
in the subsequent stages.

\textbf{Demosaicking.} 
If the input is mosaicked, we demosaick the input at this stage.
To our knowledge, most cameras use more
advanced algorithms, such as the Adaptive Homogeneity-Directed (AHD) algorithm \cite{hirakawa2005adaptive}. We
provide a Python adaptation of the reference AHD algorithm and an algorithm
developed by Kodak Inc.
\cite{hibbard1995apparatus} implemented in high performance language Halide
\cite{li2018differentiable}. We provide bilinear demosaicking as well because it
has been widely used in the recent camera simulation literature
\cite{brooks2018unprocessing,plotz2017benchmarking,guo2018toward}.
Hot/Dead pixel correction and white balance also occur here.

\textbf{Cellphone Denoising.} Demosaicking noisy images tends to result in
long-grained artifacts (as
Fig. \ref{fig:awgn_vs_real_noise} shows). Our third stage applies denoising to the image. We include 
three denoising algorithms--bilateral filters \cite{tomasi1998bilateral}, median
filters \cite{huang1979fast}, and wavelet coring \cite{donoho1994ideal}--with the
option to turn each algortihm on/off as well as reorder them. Performing tonemapping prior to denoising can be beneficial for
denoising different intensity ranges because it can non-linearly compress
a particular range of the intensity leading to a more smoothing effect. We include a pre-tonemapping operator,
which can be a gamma or an s-shape tone curve.

\textbf{Tonemapping and Post-processing.} The last stage performs postprocessing
that aims to generally improve the aesthetics of the
image. We include saturation adjustment, tonemapping for additional
tone/contrast enhancement, unsharp mask for detail enhancement, and JPEG
compression for JPEG compression artifacts simulation.

We build our pipeline largely on top of the PyTorch package
\cite{paszke2017automatic}. This allows us to readily integrate it
into learning frameworks. Because some of the software used does not support
differentiation \cite{scikit_image}, we do not utilize the differentiability of
our pipeline.
While we believe that our pipeline is realistic and rich in features, it is by no means a
comprehensive set of operations implemented by camera manufacturers. In particular, we
do not consider automatic adjustments such as auto-exposure and auto
white-balance. These modules will become crucial in automatic processing of
cellphone images.
Nonetheless, we demonstrate in section \ref{sec:eval_pipe} that our pipeline
can emulate cellphone processing, given an appropriate set of
parameters.

\subsection{Camera Simulation Evaluation}
\label{sec:eval_pipe}

We show that our pipeline is expressive enough to perform the same image processing as a camera's image signal
processing (ISP) unit. Fortunately, modern cellphones allow RAW and JPEG captures
from the same exposure. This means that if we are able to process the RAW image
into the same, or similar, JPEG image as captured, we will have successfully
emulated the camera's ISP.

Because our pipeline is missing automatic adjustments commonly found in a camera ISP, we
allow adjustments of parameters to individual images. In particular, tones
and color balance are adaptively adjusted per image/scene.
Denoising parameters, on the other hand, are held fixed per camera to reduce
the risk of overfitting.

We captured RAW + JPEG images with an iPhone
7, an iPhone 8, and a Samsung Galaxy S7. We chose these phones because they
are recent enough to allow the capture of RAW but not too recent as to have
superior imaging sensors. Including both iOS and Android phones demonstrates the versatility of our pipeline because they are
likely to have different processings and imaging sensors. We captured
approximately 10 scenes on each phone. We focus on low-light scenes so that the
noise pattern is visible, allowing us to evaluate the similarity of the
processing results of our pipeline.

To find the best parameters for each image, we
performed grid search of tone and color parameters, using L2 loss on the luminance
and chrominance channels respectively. We then hand-tuned each parameter to
obtain the final result.

\subsection{Evaluation Result}
\label{sec:iphone8}

\begin{figure}[t]
\centering
    \begin{tabularx}{\linewidth}{*{3}{Y}}
     Real JPEG &  & Simulated \\
     \multicolumn{3}{c}{\includegraphics[width=1\linewidth]{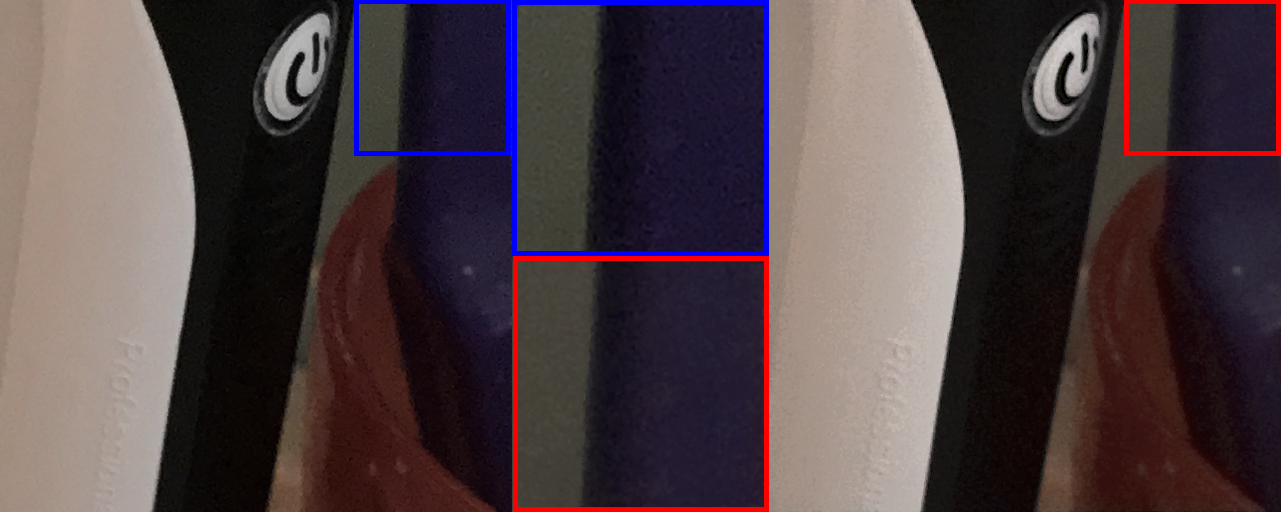}} \\
     \multicolumn{3}{c}{iPhone 7 (30.6 dB)} \\
     \multicolumn{3}{c}{\includegraphics[width=1\linewidth]{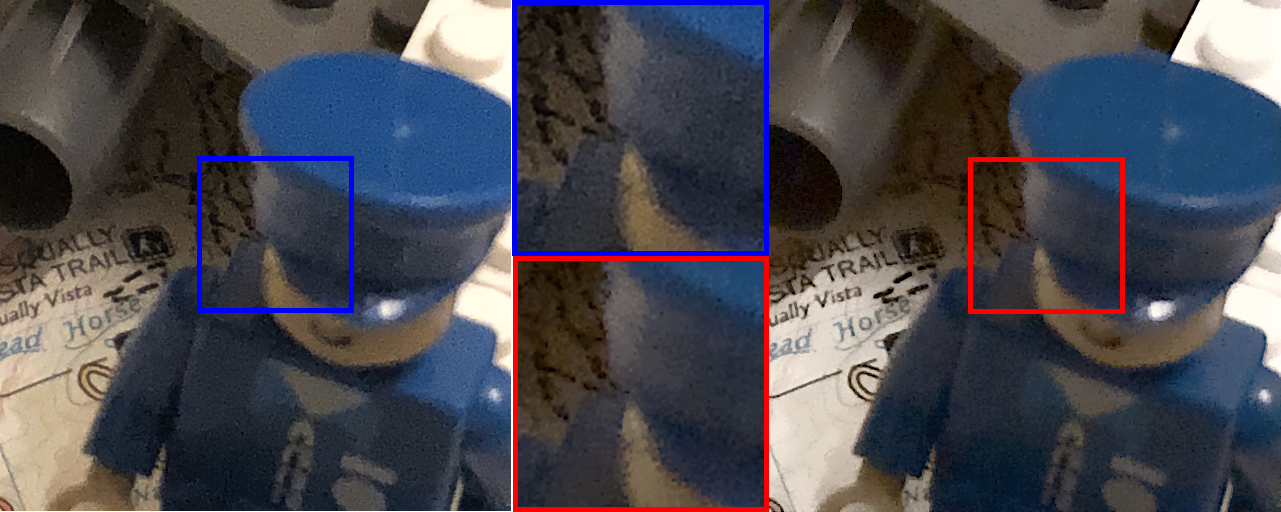}} \\
     \multicolumn{3}{c}{iPhone 8 (25.8 dB)} \\
     \multicolumn{3}{c}{\includegraphics[width=1\linewidth]{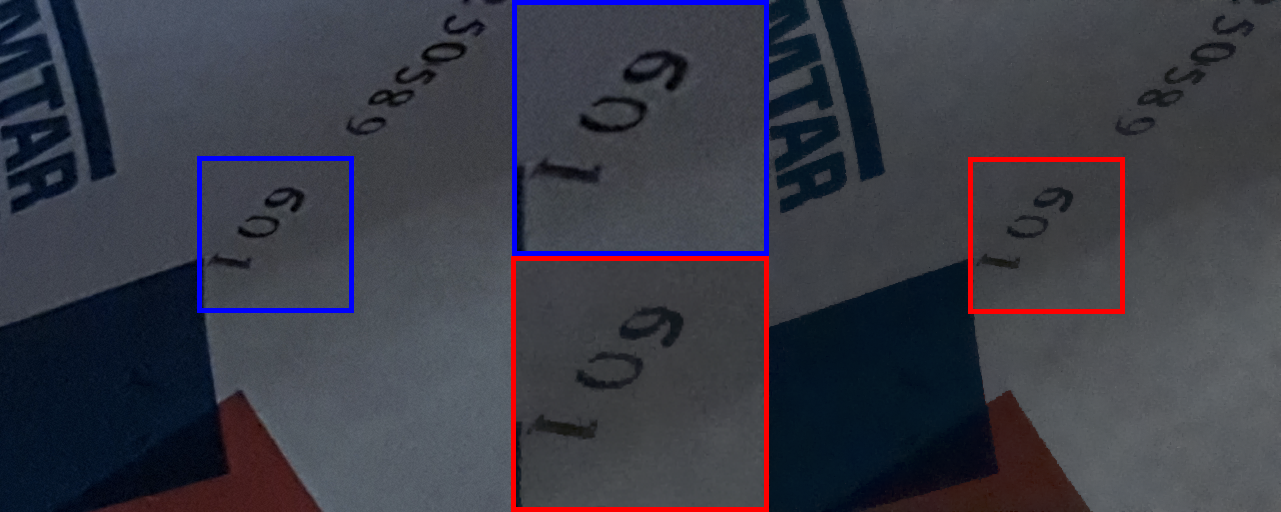}} \\
     \multicolumn{3}{c}{Samsung Galaxy S7 (32.2 dB)} \\

 \end{tabularx}
    \caption{
    Comparison of our processed RAW and camera JPEG for iPhone 7, iPhone 8, and Samsung
    Galaxy S7. The tone and noise pattern match well. 
    For more results, please refer to the supplementary material.
}
\label{fig:raw_v_jpg}
\end{figure}

Fig. \ref{fig:raw_v_jpg} shows the comparison of our pipeline processing and the
camera JPEG. Our simulation obtains an average PSNR of 28.9dB - 30.8dB and
SSIM of 0.873 - 0.888 across the three phones. In addition to these metrics, we
visually inspect the noise pattern in both the camera JPEG and
our processed RAW, and we find them to be subjectively similar.

\begin{figure}[t]
\centering
\begin{tabularx}{\linewidth}{*{3}{Y}}
     Real JPEG &  & Simulated \\
     \multicolumn{3}{c}{\includegraphics[width=1\linewidth]{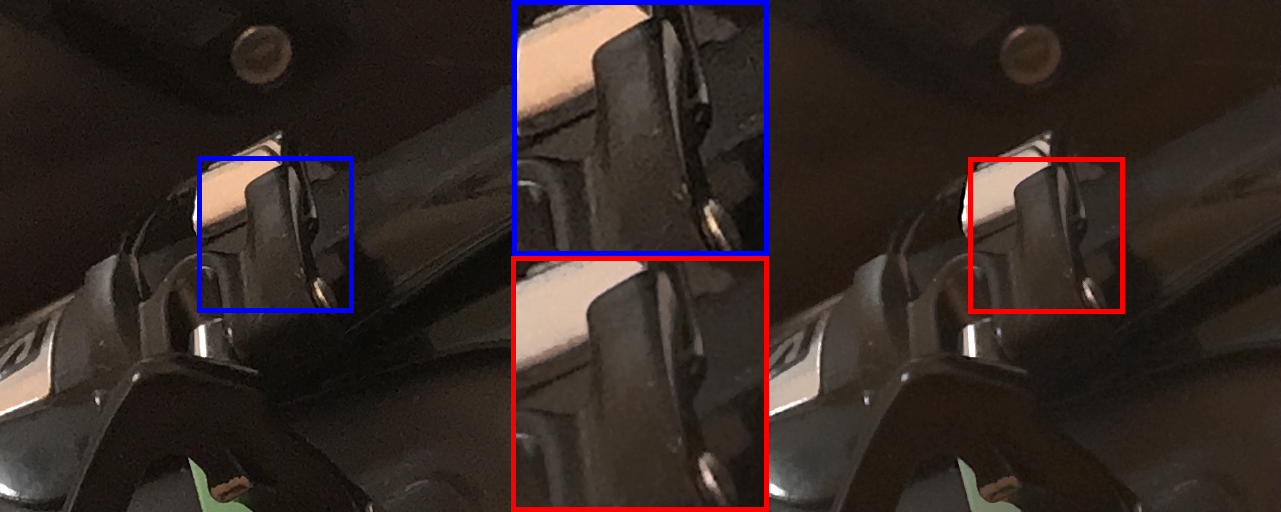}} \\
     \multicolumn{3}{c}{iPhone 7 (31.5 dB)} \\
     \multicolumn{3}{c}{\includegraphics[width=1\linewidth]{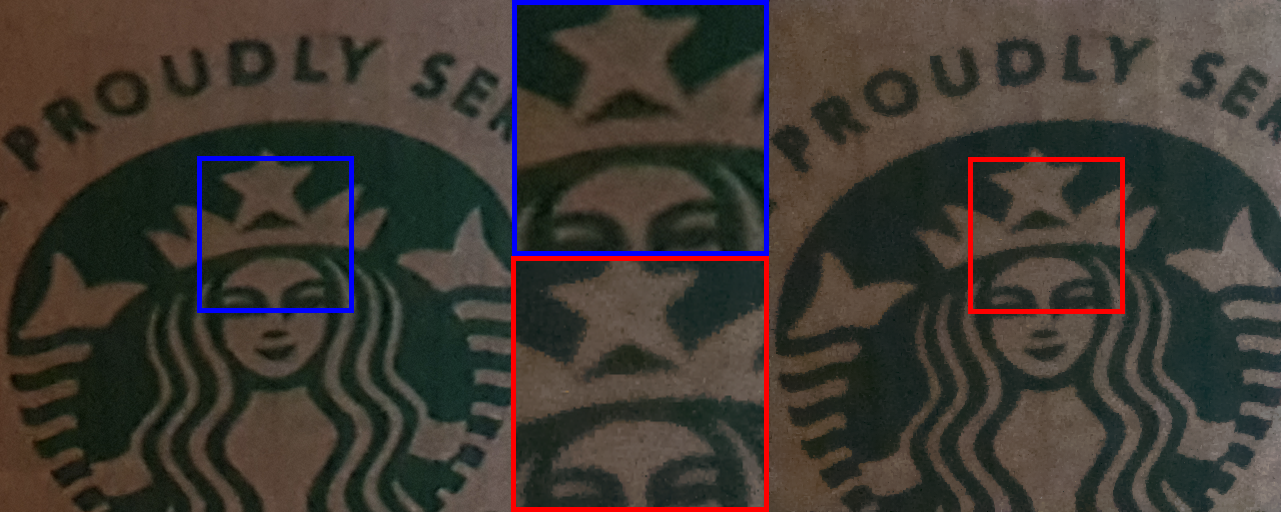}} \\
     \multicolumn{3}{c}{Samsung Galaxy S7 (26.0 dB)} \\

 \end{tabularx}

    \caption{
    Some examples where we do not match the appearance well. 
    The PSNR for the top picture is high because while we are able
    to match the tone well, but the noise pattern is over-smoothed.
}
\label{fig:raw_v_jpg_bad}
\end{figure}

While our pipeline can achieve good PSNR and SSIM numbers, these metrics tend to
over-emphasize tones and low-frequency image content. We find some visible differences in the
level of smoothing across intensity levels that may require per-image denoising
parameter tuning to remove (see Fig. \ref{fig:raw_v_jpg_bad}). Nonetheless, the level of smoothing is satisfacory overall, and we show that
this pipeline can be used to improve end-to-end denoising task in Section
\ref{sec:denoising_exp}.

\section{Denoising Experiment}
\label{sec:denoising_exp}

We demonstrate that our pipeline can be used for generating training data for
real image denoising.
Denoising is a well-studied topic, yet, few works have attempted to
model realistic noise correlation. We show that the lack of realistic noise can be
detrimental to denoising performance.

We synthetically generate our datasets using the camera simulation pipeline
described in Section \ref{sec:cam_sys}.
Using different sets of
parameters, we seek to answer two important questions: does having a realistic
noise model matter, and if so, how realistic does the noise model have to be?

Many works on denoising are shifting towards denoising RAW images, where noise is
easier to model \cite{gharbi2016deep,brooks2018unprocessing,chen2018learning}.
We focus on denoising JPEG images for two reasons. First, most image
reconstruction algorithms deal primarily with JPEG images. But for these methods, using an
additive white Gaussian noise model with JPEG images can lead to inferior results
\cite{aittala2018burst} . Second, many
photographs taken are in JPEG format because it is
often easier to work with and uses less storage. Therefore, any algorithm that
aims to be widely adopted must be able to deal with the degradation present in the
JPEG images.

We primarily focus on learning-based approaches, for which synthetic data
generation is useful for training. Methods that do not require training
data may still find it useful to generate realistic test data as an alternative to
collecting their own dataset.

\subsection{Training Data and Architecture}

\begin{figure*}[t]
\centering
\includegraphics[width=0.8\linewidth]{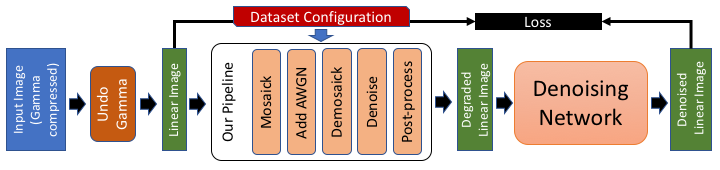}
    \caption{
        Our training setup. We use our pipeline described in Section
        \ref{sec:cam_sys} to generate the data for the denoising network. By
        varying configuration of the pipeline, datasets that simulate different
        noise profiles (AWGN, processed JPEG from cellphones, etc) can be
        generated, allowing a comparative study of these noise profiles and
        their effectiveness.
    }
        \label{fig:training}
\end{figure*}

Our denoising setup aims to denoise RGB images that have been processed by the
camera. Fig. \ref{fig:training} shows our training setup.
It starts from an input JPEG image with gamma compression. We undo
the gamma compression to obtain a linear image to be degraded by our camera simulation
pipeline (Section \ref{sec:cam_sys}). The degraded output is then fed into a
denoising CNN. Finally, the denoised image is compared to the
original linear, clean RGB image to provide training signal to the denoising
network.

\textbf{Choice of Modules in  the Camera Pipeline Simulation.} Since we
focus only on the noise pattern, we turn off all tonemapping and color
operations. This way the denoising network does not have to learn to adjust
tones, simplifying the learning problem considerably.

We observe that real cellphone denoising is often a combination of bilateral
and median filters, so we use these two algorithms in our
cellphone simulation pipeline. We find that the Kodak algorithm
\cite{hibbard1995apparatus} and AHD
\cite{hirakawa2005adaptive} perform roughly the same, so we choose
the Kodak algorithm for which we have a more efficient implementation.

\textbf{Parameters of the Pipeline.} We set the configuration of our processing pipeline based on the range of values
observed during our experiment (see Section \ref{sec:iphone8}). For
simplicity, we
randomize each parameter independently. We choose noise strengths based on
measurement data from the iPhone 7 and the Samsung Galaxy S6 at various ISO
\cite{iphone7data,samsungs6data}.  We exaggerate the noise strength to ensure
that the network sees very noisy samples in the training set. Table \ref{tab:data_configs}
lists noise strengths and processing performed on each of our datasets.

\begin{table}[h]
\centering
    \caption{Configuration of Training Datasets}
    \label{tab:data_configs}
 \small
    \begin{tabular}{|c|c|c|c|}
    \hline
        \textbf{Training Data} &
        \textbf{Gaussian STD} &
        \thead{\textbf{Poisson} \\ \textbf{Mult Factor}}&
        \thead{\textbf{Additional} \\ \textbf{Processing}}   \\
                %& Training Data & \thead{No Post-\\processing}& \thead{No \\ Denoising} & \thead{No \\Demosaicking}   \\
    \hline
     AWGN       &   0 - 0.2      &     0              &     None    \\
    \hline
    \thead{Add-Mult \\ WGN}      &   0 - 0.1      &    0 - 0.02        &     None     \\
    \hline
     Ours       &   0 - 0.1      & 0 - 0.02        & \thead{Demosaicking, \\ Denoising, \\ Post-processing} \\
    \hline
    \thead{Samsung S7 \\ Measurement \\ @ ISO800}       &   0.007      &  0.02        & N/A \\
    \hline
 \end{tabular}
\end{table}

\textbf{Source Dataset.} We use the MIT-Adobe5k
dataset \cite{bychkovsky2011learning} as our input images because it has
high-quality photographs. We use their expert-C
retouched images so that the input and target tones are representative of
JPEG images. We
downsample the images by 4x to reduce any remaining noise and artifacts.
We extract 5 patches randomly from each image in the dataset, resulting
in a total of 25k patches available for training.

\textbf{Denoising Network.} Since the focus of this work is not the network architecture, we used the author's
implementation of the Neural Nearest Neighbor network \cite{plotz2018neural}, which has been shown to
achieve state of the art result in denoising. We follow the
author's training method, using the Adam optimizer \cite{kingma2014adam} with
learning rate of 0.001. The author also noted that increasing learning rate decay
is beneficial, so we decay the learning rate by $10^{-5}$ over 100 epochs
(instead of $10^{-5}$ over 50 epochs in the original paper) and train for 100 epochs. 

\textbf{Performance Consideration.} Because our dataset is synthetic, we are
able to generate it on-the-fly. This allows us to rapidly prototype
and change configurations without pre-generating the entire dataset.
Additionally, each input patch
receives different randomized processing parameters in each epoch, which increases the complexity of our
dataset. 
Our pipeline implementation is based largely on PyTorch modules
\cite{paszke2017automatic} and uses the
high performance Halide language \cite{li2018differentiable}. While
performance varies with the system and configuration, we are able to largely
saturate a machine with a Tesla P100 GPU and 32 CPU cores (80x80 patch, batch size=32).
Training takes roughly 9 hours.

\subsection{Testing Data}

Because we focus on denoising real JPEG photographs, real
JPEG images are required to measure the denoising performance.
This is challenging because we do not have access to the
blackbox camera processing, and our pipeline cannot process large
amounts of images automatically. Furthermore, some artifacts in the JPEG images cannot be
removed by averaging.

Existing datasets do not provide the required clean JPEG images.
\cite{abdelhamed2018high} and \cite{chen2018learning} provide only RAW images,
while \cite{plotz2017benchmarking} uses simple processing which may not be
realistic. \cite{schwartz2019deepisp} provides short- and long-exposure image
pairs, but they do not keep exposure levels constant, resulting in large tone shifts between the
ground truth and noisy images. Furthermore, we find the noise in
their long-exposure ground truth images to still be significant. 

Because of these limitations, we use averaged RAW images from bursts as the
target. Noise in RAW images is zero-mean and can be reduced by averaging. 
However, using RAW images as the target requires demosaicking and normalizing the tone. Because PSNR and SSIM are very sensitive
to tone change, we normalize each ground truth image to the output image at test
time by matching their means and standard deviations per color channel.

We collected test images using the iPhone 8, Pixel XL, and Samsung Galaxy S7 to
test generalization across camera models. For each phone, roughly 20-25 scenes were
captured,
and for each scene, one high-ISO image and a burst of 10 low-ISO 
images were taken. All images were captured in the RAW + JPEG format and the
exposure were kept roughly constant. We used
sturdy tripods and avoided moving objects and reflections as much as possible. We
also set a timer and used a shutter cable to avoid any movement that resulted from
interacting with the phones.

\section{Denoising Results}

In this Section, we report the findings of our denoising experiments.

\subsection{Additive/Multiplicative White Gaussian Noise (AMWGN) vs Realistic Noise}

% sample table
%\begin{table*}
%\centering
 %\caption{Mean IoU of our modified FusionNet architecture.}
 %\label{accuracy-table}
 %\small
 %\begin{tabular}{l|lllll|lll|lll}
   %\toprule
   %Training  & \multicolumn{5}{|c|}{Single quality} & \multicolumn{3}{|c}{Coarse + 1k fine} & \multicolumn{3}{|c}{Coarse + 3k fine} \\
   %\midrule
   %Num images  & 3k & 6k & 12k & 24k & 48k & 12k+1k & 24k+1k & 48k+1k & 12k+3k & 24k+3k & 48k+3k \\
   %\midrule
     %fine--labels & 85.26 & 88.75 & 89.61 & 91.89 & \textbf{93.93} &  n/a  &  n/a  &  n/a  & n/a   &  n/a  &  n/a  \\
     %4px error    & 84.97 & 86.93 & 90.06 & 91.56 & 93.33          & 90.58 & 92.11 & \textbf{94.02} & 90.61 & 92.49 & \textbf{94.31}  \\
     %8px error    & 84.20 & 86.20 & 88.61 & 90.58 & 92.01          & 90.17 & 91.95 & 93.65 & 90.62 & 91.42 & 93.64 \\
     %16px error   & 82.15 & 83.97 & 86.92 & 88.42 & 90.65          & 89.60 & 90.44 & 92.38 & 89.81 & 90.75 & 92.96 \\
     %32px error   & 78.30 & 81.96 & 83.57 & 86.53 & 88.42          & 89.41 & 91.26 & 91.41 & 89.32 & 90.08 & 92.03 \\
   %\bottomrule
 %\end{tabular}
%\end{table*}

\begin{table}
\centering
 \caption{Quantitative comparison of training data between AWGN vs our pipeline for different cellphone cameras.}
 \label{tab:awgn_v_pipeline}
 \small
 \begin{tabular}{|c|c|ccc|}
     \hline
      \multirow{2}{*}{Metric}  & \multirow{2}{*}{\thead{Input vs\\ Ground truth}}  & \multicolumn{3}{c|}{Training Data}  \\
     \cline{3-5}
                   &        & AWGN   & AMWGN & Our Pipeline   \\
   \hline
      \multicolumn{5}{|c|}{iPhone 8} \\
   \hline
     PSNR       &  32.4 & 32.3    &     32.6              &  \textbf{35.2}   \\
     SSIM       & 0.788 & 0.788   &    0.799              &  \textbf{0.892}  \\
   \hline
      \multicolumn{5}{|c|}{Pixel XL} \\
   \hline
     PSNR       &  31.2 & 31.3    &     31.6              &  \textbf{35.1}   \\
     SSIM       & 0.760 & 0.761   &    0.775              &  \textbf{0.881}  \\
   \hline
      \multicolumn{5}{|c|}{Samsung Galaxy S7} \\
   \hline
     PSNR       &  41.1 & 41.0    &     41.4              &  \textbf{42.5}   \\
     SSIM       & 0.933 & 0.931   &    0.939              &  \textbf{0.954}  \\
    \hline
 \end{tabular}
\end{table}

The network trained on our dataset significantly outperforms ones that were
trained with additive/multiplicative Gaussian noise. Table
\ref{tab:awgn_v_pipeline} shows denoising results of N3Net
\cite{plotz2018neural} trained with different datasets. 
On the iPhone 8 and Pixel XL test sets, the model trained on our dataset
achieved a 3 dB higher PSNR and nearly 0.1 higher SSIM.
On the Samsung Galaxy S7, the improvement is approximately 1.5 dB in PSNR and
0.015 in SSIM. These are significant margins because many recent denoising works
often report improvements that are less than 0.5-1 dB
\cite{plotz2018neural,zhang2017beyond}.

\begin{figure*}[t]
\centering
\begin{tabular}{ccccc}
    Ground Truth & Noisy Input & AWGN & AMGN & Ours \\
    \midrule

        \includegraphics[width=0.15\linewidth]{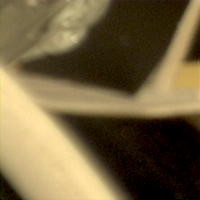} &
        \includegraphics[width=0.15\linewidth]{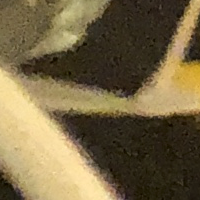} &
        \includegraphics[width=0.15\linewidth]{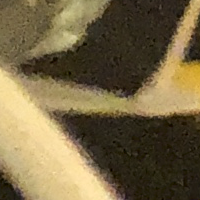} &
        \includegraphics[width=0.15\linewidth]{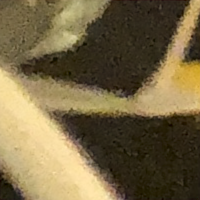} &
        \includegraphics[width=0.15\linewidth]{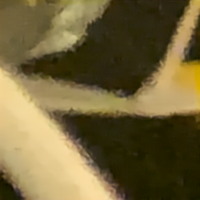} \\

        \includegraphics[width=0.15\linewidth]{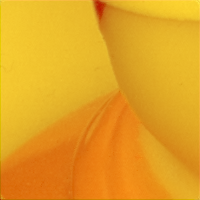} &
        \includegraphics[width=0.15\linewidth]{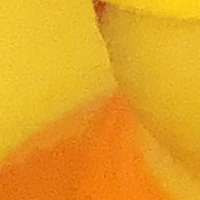} &
        \includegraphics[width=0.15\linewidth]{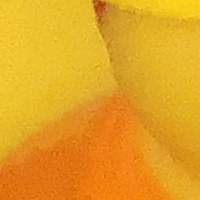} &
        \includegraphics[width=0.15\linewidth]{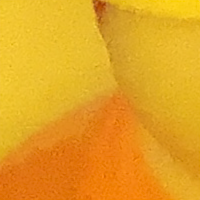} &
        \includegraphics[width=0.15\linewidth]{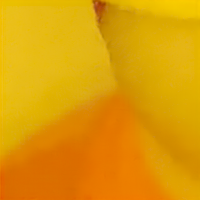} \\
        \multicolumn{5}{c}{iPhone 8} \\

    \midrule

        \includegraphics[width=0.15\linewidth]{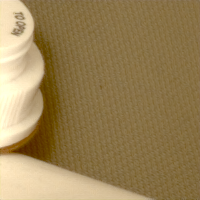} &
        \includegraphics[width=0.15\linewidth]{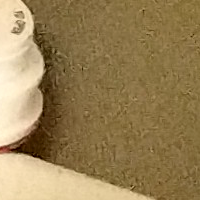} &
        \includegraphics[width=0.15\linewidth]{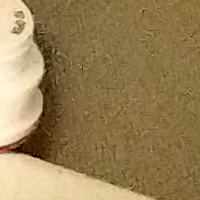} &
        \includegraphics[width=0.15\linewidth]{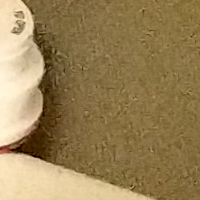} &
        \includegraphics[width=0.15\linewidth]{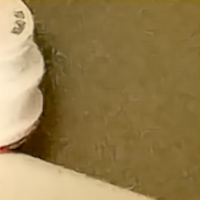} \\

        \includegraphics[width=0.15\linewidth]{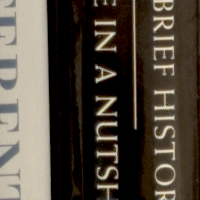} &
        \includegraphics[width=0.15\linewidth]{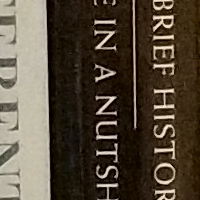} &
        \includegraphics[width=0.15\linewidth]{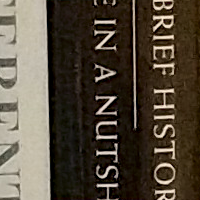} &
        \includegraphics[width=0.15\linewidth]{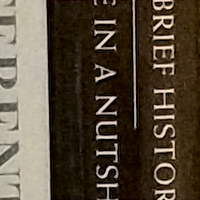} &
        \includegraphics[width=0.15\linewidth]{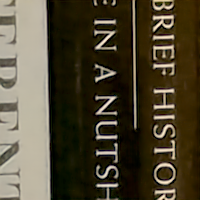} \\
        \multicolumn{5}{c}{Pixel XL} \\

    \midrule
        \includegraphics[width=0.15\linewidth]{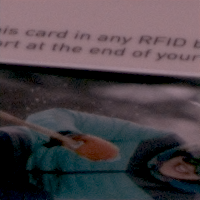} &
        \includegraphics[width=0.15\linewidth]{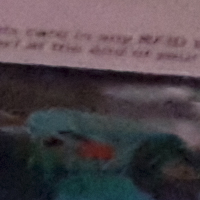} &
        \includegraphics[width=0.15\linewidth]{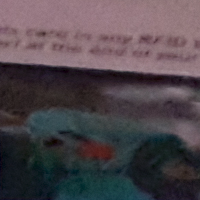} &
        \includegraphics[width=0.15\linewidth]{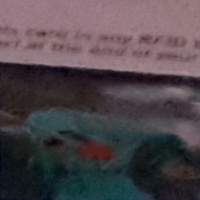} &
        \includegraphics[width=0.15\linewidth]{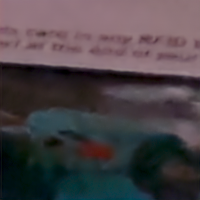} \\

        \includegraphics[width=0.15\linewidth]{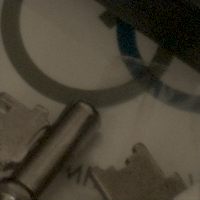} &
        \includegraphics[width=0.15\linewidth]{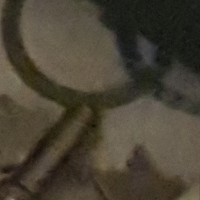} &
        \includegraphics[width=0.15\linewidth]{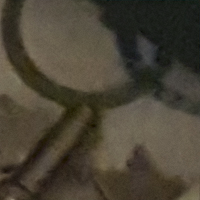} &
        \includegraphics[width=0.15\linewidth]{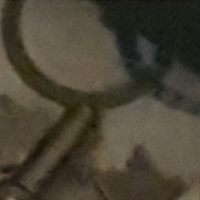} &
        \includegraphics[width=0.15\linewidth]{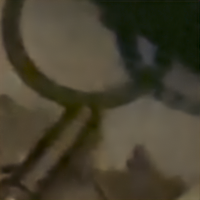} \\
        \multicolumn{5}{c}{Samsung Galaxy S7} \\

    \midrule

 \end{tabular}    
    \caption{
 Sample output patches from denoising networks trained with AWGN vs our dataset,
 on iPhone 8, Pixel XL, and Samsung Galaxy S7 test data. More results in the
 supplementary material.
}
\label{fig:awgn_v_pipeline}
\end{figure*}

Visual inspection of the resulting denoised patch reveals that the AMWGN models
seem to ignore noise entirely--the output patch is almost identical to the
input patch, as Fig. \ref{fig:awgn_v_pipeline} shows. On the iPhone 8 test data,
the PSNR between the input
and output patches are over 50dB, and the SSIM is over 0.996 (vs 35.7 dB and
0.856 for our model). 
% The A/MWGN models failed to denoise real images.

\begin{figure}[h]
\centering
\setlength\tabcolsep{1.5pt}
\begin{tabular}{ccc}
    \toprule
    \thead{Input w/ AWGN added} & \thead{Denoised by \\AWGN Network} & \thead{Denoised by
    \\A\textbf{M}WGN Network} \\
    \midrule
        %\multicolumn{6}{c}{iPhone 8} \\
        \includegraphics[width=0.33\linewidth]{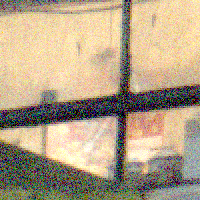} &
        \includegraphics[width=0.33\linewidth]{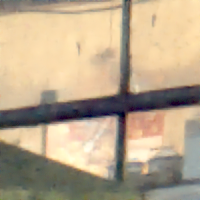} &
        \includegraphics[width=0.33\linewidth]{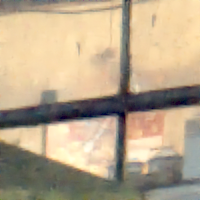} \\

     \bottomrule
 \end{tabular}
    \caption{
        Our AWGN-trained and Add-Mult WGN models are able to denoise patches with AWGN, suggesting
        that the models are working correctly. More result is in the supplementary material.
    }
\label{fig:debug_awgn}
\end{figure}

To show that our AWGN model works correctly, we pass the patches with
additive Gaussian noise with STD of 0.1 (on a 0-1 scale) to the model. Fig. \ref{fig:debug_awgn}
 shows the denoising results. The AWGN networks are able to properly denoise the patches
with PSNRs of 36.6 dB and 36.0 dB for the additive and additive-multiplicative
models, respectively. This suggests that their performance on real images is likely
the result of a mismatch between real test JPEG image and the additive Gaussian
noise training data, and not the faulty implementation of our models.

\textbf{Denoising Demosaicked RAW.} Most image reconstruction algorithms are designed for RGB
images, so when working with RAW images, demosaicking is often applied (except
for a few works \cite{gharbi2016deep,heide2014flexisp}). We
demosaick our real RAW noisy images and use them as test input. We find
that our data outperforms AWGN by 7-9 dB in PSNR and 0.2-0.3 on SSIM,
depending on the demosaicking algorithms applied (the training always uses
Kodak \cite{hibbard1995apparatus}).

\subsection{Ablation Study}

In order to understand the essential features of our pipeline, we train
additional networks with different components of our pipeline turned off. We group the
components based on stages outlined in Section \ref{sec:cam_sys}: demosaicking,
denoising, and post-processing.

\begin{table}[h]
\centering
    \caption{Quantitative comparison of handicapped data generation by turning
    one component off at a time  (iPhone 8 test data).}
    \label{tab:ablation1}
 \small
    \begin{tabular}{|c|cccc|}
    \hline
     Metric         & Full-Pipeline & \thead{No Post-\\processing}& \thead{No \\ Denoising} & \thead{No \\Demosaicking}   \\
    \hline
     PSNR       &  35.2         & 34.0              &     34.7     &  33.9 \\
     SSIM       & 0.892         & 0.866             &    0.870     &  0.846 \\
    \hline
 \end{tabular}
\end{table}

\begin{figure}[h]
\centering
\setlength\tabcolsep{1.5pt}
\begin{tabular}{ccc}
    \toprule
        No Post-Processing & No Denoising & No Demosaicking \\
    \midrule
        %\multicolumn{6}{c}{iPhone 8} \\

        %\includegraphics[width=0.33\linewidth]{figs/results/iphone8_0318_iphone7_extra/img659.png} &
        %\includegraphics[width=0.33\linewidth]{figs/results/iphone8_0318_iphone7_extra/img660.png} &
        %\includegraphics[width=0.33\linewidth]{figs/results/iphone8_0318_iphone7_extra/img661.png} \\

        \includegraphics[width=0.33\linewidth]{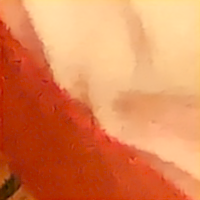} &
        \includegraphics[width=0.33\linewidth]{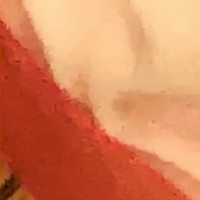} &
        \includegraphics[width=0.33\linewidth]{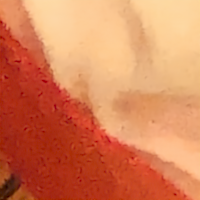} \\

     \bottomrule
 \end{tabular}
    \caption{
        Qualitative evaluation of our ablation datasets (on the iPhone 8 test data). The network
        trained without post-processing is able to smooth the real image,
        while the network without denoising and demosaicking shows less
        smoothing of the noise.
    }
    \vskip -5mm
\label{fig:remove_comp}
\end{figure}

\textbf{Denoising and Demosaicking.} We find demosaicking and denoising to be
important to the smoothing of the image. Fig. \ref{fig:remove_comp} shows a
sample patch from three different networks: ones that are trained without
post-processing, without denoising, and without demosaicking. The network
trained without post-processing produces the smoothest outputs, while the other two retain
long-grained artifacts present in the input image. Table \ref{tab:ablation1} shows the
quantitative result for these networks. While the
PSNRs are comparable, removing demosaicking suffers the largest reduction in
SSIM, confirming our qualitative observation. For brevity, we only show results on iPhone 8
test data, but we observe similar trends on both Pixel XL and Samsung Galaxy S7
test data.

\begin{table}[h]
\centering
    \caption{Quantitative comparison on the choice of demosaicking algorithm (iPhone 8 test data).}
    \label{tab:ablation2}
 \small
    \begin{tabular}{|c|cccc|}
    \hline
         Metric & \thead{Full \\ Pipeline} & \thead{Kodak \cite{hibbard1995apparatus}}&  AHD\cite{hirakawa2005adaptive} &  Bilinear  \\
    \hline
     PSNR       &  35.2         & 33.6              &     32.9     &  31.1 \\
     SSIM       & 0.892         & 0.840             &    0.821     &  0.746 \\
    \hline
 \end{tabular}
\end{table}

\begin{figure}[h]
\centering
\setlength\tabcolsep{1.5pt}
\begin{tabular}{ccc}
    \toprule
        Kodak \cite{hibbard1995apparatus}& AHD\cite{hirakawa2005adaptive} &  Bilinear \\
    \midrule
        %\multicolumn{6}{c}{iPhone 8} \\

        %\includegraphics[width=0.33\linewidth]{figs/results/iphone8_0318_iphone7_demosaick/img9.png} &
        %\includegraphics[width=0.33\linewidth]{figs/results/iphone8_0318_iphone7_demosaick/img10.png} &
        %\includegraphics[width=0.33\linewidth]{figs/results/iphone8_0318_iphone7_demosaick/img11.png} \\

        \includegraphics[width=0.33\linewidth]{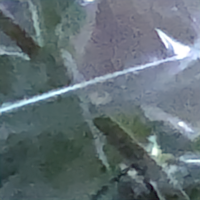} &
        \includegraphics[width=0.33\linewidth]{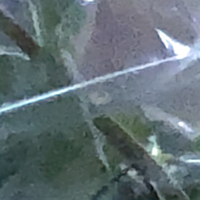} &
        \includegraphics[width=0.33\linewidth]{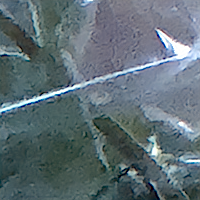} \\

     \bottomrule

 \end{tabular}
    \caption{
        Qualitative evaluation of simulation with different demosaicking
        algorithms  (iPhone 8 test data). The networks
        trained with edge-aware demosaicking (Kodak \cite{hibbard1995apparatus},
        AHD \cite{hirakawa2005adaptive}) were able to smooth images well, while
        one with the commonly used bilinear demosaicking retains the most
        artifacts.
        More result is in the supplementary material.
    }
    \vskip -5mm
\label{fig:demosaick_choice}
\end{figure}

\textbf{Choice of Demosaicking Algorithm.} We further investigate the choice of
demosaciking algorithm used, because most works that simulate the camera processing pipeline
use bilinear demosaicking
\cite{brooks2018unprocessing,guo2018toward,plotz2017benchmarking}.

As Fig. \ref{fig:demosaick_choice} shows, network trained with bilinear interpolation 
retains the most JPEG artifacts in their denoising result.
On the other hand, the networks trained with the other two edge-aware
demosaicking algorithms are
able to remove more of these artifacts. Table \ref{tab:ablation2} shows the
quantitative results. The AHD \cite{hirakawa2005adaptive} and the Kodak
algorithm \cite{hibbard1995apparatus} outperforms bilinear demosaicking by more than 2dB on
PSNR and over 0.07 on SSIM.

%\textbf{Parameters in Section \ref{sec:eval_pipe}} \tiam{this could be in
%limitation section.} One might expect that the denoising parameters that allow
%us to simulate processing of camera (found in section \ref{sec:eval_pipe}) to
%be applicable to our training. However, because the input to the pipelines
%during training are already tone-mapped, this have significant, and non-trivial
%effects on the appearance of the training data.

\section{Conclusion and Future Work}

We have proposed a realistic camera pipeline simulation that is expressive
enough to process RAW inputs into JPEG images that is visually similar to the ones cameras
produce. We use this simulation to generate realistic datasets for training
denoising CNNs and show that it improves the performance of such networks on real
JPEG images by over 3dB. Demosaicking and denoising seem to be the most
important components of our pipeline that enable such improvement. Removing
either of them leads to a significant drop in the quality of the denoised output.
Using correct algorithms for these components is also important. The bilinear
demosaicking algorithm commonly used in previous camera simulation work
\cite{brooks2018unprocessing,guo2018toward,plotz2017benchmarking} leads to
a significant performance drop, while edge-aware algorithms such as AHD
\cite{hirakawa2005adaptive} do not.

While we have shown our pipeline is useful and realistic, it still requires
significant manual tuning in order to match the appearance of the processed JPEG. The
ability to automatically match the appearance is an interesting future
direction. This will help ensure realism, so that the generated data can be used
for any arbitrary camera models.

\section*{Acknowledgments}

The authors would like to thank the Toyota Research Institute
for their generous support of the projects. We thank Tzu-Mao Li for his
helpful comments, and Luke Anderson for his help revising this draft.

{\small
    \bibliographystyle{ieee}
    \bibliography{phone2dslr_reference}

\begin{thebibliography}{10}\itemsep=-1pt

\bibitem{iphone7data}
Read noise in dns versus iso setting.
\newblock
  \url{http://www.photonstophotos.net/Charts/RN_ADU.htm\#Apple\%20iPhone\%207\_12,Samsung\%20Galaxy\%20S6(S5K2P2)\_10}.
\newblock Accessed: 2018-10-30.

\bibitem{samsungs6data}
Samsung galaxy s6 edge : Measurements - dxomark.
\newblock
  \url{https://www.dxomark.com/Cameras/Samsung/Galaxy-S6-Edge---Measurements}.
\newblock Accessed: 2018-10-30.

\bibitem{abdelhamed2018high}
A.~Abdelhamed, S.~Lin, and M.~S. Brown.
\newblock A high-quality denoising dataset for smartphone cameras.
\newblock In {\em Proceedings of the IEEE Conference on Computer Vision and
  Pattern Recognition}, pages 1692--1700, 2018.

\bibitem{aittala2018burst}
M.~Aittala and F.~Durand.
\newblock Burst image deblurring using permutation invariant convolutional
  neural networks.
\newblock In {\em Proceedings of the European Conference on Computer Vision
  (ECCV)}, pages 731--747, 2018.

\bibitem{brooks2018unprocessing}
T.~Brooks, B.~Mildenhall, T.~Xue, J.~Chen, D.~Sharlet, and J.~T. Barron.
\newblock Unprocessing images for learned raw denoising.
\newblock {\em arXiv preprint arXiv:1811.11127}, 2018.

\bibitem{bychkovsky2011learning}
V.~Bychkovsky, S.~Paris, E.~Chan, and F.~Durand.
\newblock Learning photographic global tonal adjustment with a database of
  input/output image pairs.
\newblock In {\em CVPR 2011}, pages 97--104. IEEE, 2011.

\bibitem{chen2018learning}
C.~Chen, Q.~Chen, J.~Xu, and V.~Koltun.
\newblock Learning to see in the dark.
\newblock In {\em Proceedings of the IEEE Conference on Computer Vision and
  Pattern Recognition}, pages 3291--3300, 2018.

\bibitem{dabov2006image}
K.~Dabov, A.~Foi, V.~Katkovnik, and K.~Egiazarian.
\newblock Image denoising with block-matching and 3d filtering.
\newblock In {\em Image Processing: Algorithms and Systems, Neural Networks,
  and Machine Learning}, volume 6064, page 606414. International Society for
  Optics and Photonics, 2006.

\bibitem{dong2018learning}
J.~Dong, J.~Pan, D.~Sun, Z.~Su, and M.-H. Yang.
\newblock Learning data terms for non-blind deblurring.
\newblock In {\em Proceedings of the European Conference on Computer Vision
  (ECCV)}, pages 748--763, 2018.

\bibitem{donoho1994ideal}
D.~L. Donoho and J.~M. Johnstone.
\newblock Ideal spatial adaptation by wavelet shrinkage.
\newblock {\em biometrika}, 81(3):425--455, 1994.

\bibitem{farrell2003simulation}
J.~E. Farrell, F.~Xiao, P.~B. Catrysse, and B.~A. Wandell.
\newblock A simulation tool for evaluating digital camera image quality.
\newblock In {\em Image Quality and System Performance}, volume 5294, pages
  124--132. International Society for Optics and Photonics, 2003.

\bibitem{gharbi2016deep}
M.~Gharbi, G.~Chaurasia, S.~Paris, and F.~Durand.
\newblock Deep joint demosaicking and denoising.
\newblock {\em ACM Transactions on Graphics (TOG)}, 35(6):191, 2016.

\bibitem{guo2018toward}
S.~Guo, Z.~Yan, K.~Zhang, W.~Zuo, and L.~Zhang.
\newblock Toward convolutional blind denoising of real photographs.
\newblock {\em arXiv preprint arXiv:1807.04686}, 2018.

\bibitem{hasinoff2016burst}
S.~W. Hasinoff, D.~Sharlet, R.~Geiss, A.~Adams, J.~T. Barron, F.~Kainz,
  J.~Chen, and M.~Levoy.
\newblock Burst photography for high dynamic range and low-light imaging on
  mobile cameras.
\newblock {\em ACM Transactions on Graphics (TOG)}, 35(6):192, 2016.

\bibitem{heide2014flexisp}
F.~Heide, M.~Steinberger, Y.-T. Tsai, M.~Rouf, D.~Pajak, D.~Reddy, O.~Gallo,
  J.~Liu, W.~Heidrich, K.~Egiazarian, et~al.
\newblock Flexisp: A flexible camera image processing framework.
\newblock {\em ACM Transactions on Graphics (TOG)}, 33(6):231, 2014.

\bibitem{hibbard1995apparatus}
R.~H. Hibbard.
\newblock Apparatus and method for adaptively interpolating a full color image
  utilizing luminance gradients, Jan.~17 1995.
\newblock US Patent 5,382,976.

\bibitem{hirakawa2005adaptive}
K.~Hirakawa and T.~W. Parks.
\newblock Adaptive homogeneity-directed demosaicing algorithm.
\newblock {\em IEEE Transactions on Image Processing}, 14(3):360--369, 2005.

\bibitem{huang1979fast}
T.~Huang, G.~Yang, and G.~Tang.
\newblock A fast two-dimensional median filtering algorithm.
\newblock {\em IEEE Transactions on Acoustics, Speech, and Signal Processing},
  27(1):13--18, 1979.

\bibitem{karaimer2016software}
H.~C. Karaimer and M.~S. Brown.
\newblock A software platform for manipulating the camera imaging pipeline.
\newblock In {\em European Conference on Computer Vision}, pages 429--444.
  Springer, 2016.

\bibitem{kingma2014adam}
D.~P. Kingma and J.~Ba.
\newblock Adam: A method for stochastic optimization.
\newblock {\em arXiv preprint arXiv:1412.6980}, 2014.

\bibitem{krizhevsky2012imagenet}
A.~Krizhevsky, I.~Sutskever, and G.~E. Hinton.
\newblock Imagenet classification with deep convolutional neural networks.
\newblock In {\em Advances in neural information processing systems}, pages
  1097--1105, 2012.

\bibitem{lehtinen2018noise2noise}
J.~Lehtinen, J.~Munkberg, J.~Hasselgren, S.~Laine, T.~Karras, M.~Aittala, and
  T.~Aila.
\newblock Noise2noise: Learning image restoration without clean data.
\newblock {\em arXiv preprint arXiv:1803.04189}, 2018.

\bibitem{li2018differentiable}
T.-M. Li, M.~Gharbi, A.~Adams, F.~Durand, and J.~Ragan-Kelley.
\newblock Differentiable programming for image processing and deep learning in
  halide.
\newblock {\em ACM Transactions on Graphics (TOG)}, 37(4):139, 2018.

\bibitem{paszke2017automatic}
A.~Paszke, S.~Gross, S.~Chintala, G.~Chanan, E.~Yang, Z.~DeVito, Z.~Lin,
  A.~Desmaison, L.~Antiga, and A.~Lerer.
\newblock Automatic differentiation in pytorch.
\newblock In {\em NIPS-W}, 2017.

\bibitem{plotz2017benchmarking}
T.~Plotz and S.~Roth.
\newblock Benchmarking denoising algorithms with real photographs.
\newblock In {\em Proceedings of the IEEE Conference on Computer Vision and
  Pattern Recognition}, pages 1586--1595, 2017.

\bibitem{plotz2018neural}
T.~Pl{\"o}tz and S.~Roth.
\newblock Neural nearest neighbors networks.
\newblock In {\em Advances in Neural Information Processing Systems}, pages
  1095--1106, 2018.

\bibitem{portilla2003image}
J.~Portilla, V.~Strela, M.~J. Wainwright, and E.~P. Simoncelli.
\newblock Image denoising using scale mixtures of gaussians in the wavelet
  domain.
\newblock {\em IEEE Trans Image Processing}, 12(11), 2003.

\bibitem{schwartz2019deepisp}
E.~Schwartz, R.~Giryes, and A.~M. Bronstein.
\newblock Deepisp: Toward learning an end-to-end image processing pipeline.
\newblock {\em IEEE Transactions on Image Processing}, 28(2):912--923, 2019.

\bibitem{simoncelli1996noise}
E.~P. Simoncelli and E.~H. Adelson.
\newblock Noise removal via bayesian wavelet coring.
\newblock In {\em Proceedings of 3rd IEEE International Conference on Image
  Processing}, volume~1, pages 379--382. IEEE, 1996.

\bibitem{tomasi1998bilateral}
C.~Tomasi and R.~Manduchi.
\newblock Bilateral filtering for gray and color images.
\newblock In {\em null}, page 839. IEEE, 1998.

\bibitem{scikit_image}
S.~van~der Walt, J.~L. {S}ch\"onberger, J.~{Nunez-Iglesias}, F.~{B}oulogne,
  J.~D. {W}arner, N.~{Y}ager, E.~{G}ouillart, T.~{Y}u, and the scikit-image
  contributors.
\newblock scikit-image: image processing in {P}ython.
\newblock {\em PeerJ}, 2:e453, 6 2014.

\bibitem{zhang2017beyond}
K.~Zhang, W.~Zuo, Y.~Chen, D.~Meng, and L.~Zhang.
\newblock Beyond a gaussian denoiser: Residual learning of deep cnn for image
  denoising.
\newblock {\em IEEE Transactions on Image Processing}, 26(7):3142--3155, 2017.

\bibitem{zhang2018ffdnet}
K.~Zhang, W.~Zuo, and L.~Zhang.
\newblock Ffdnet: Toward a fast and flexible solution for cnn-based image
  denoising.
\newblock {\em IEEE Transactions on Image Processing}, 27(9):4608--4622, 2018.

\end{thebibliography}
}

\end{document}